\documentclass[sigconf, authorversion]{acmart}

\usepackage{lipsum}
\usepackage{amsthm}

\usepackage{xargs}        
\usepackage[normalem]{ulem}

\def\HiLi{\leavevmode\rlap{\hbox to 0.825\hsize{\color{gray!20}\leaders\hrule height .8\baselineskip depth .5ex\hfill}}}

\usepackage{url}

\usepackage[scientific-notation=true]{siunitx}

\usepackage{algorithm,algorithmicx,algpseudocode}
\usepackage{url}

\usepackage[normalem]{ulem}
\useunder{\uline}{\ul}{}

\usepackage[colorinlistoftodos,prependcaption,textsize=tiny]{todonotes}
\newcommandx{\unsure}[2][1=]{\todo[linecolor=red,backgroundcolor=red!25,bordercolor=red,#1]{#2}}
\newcommandx{\change}[2][1=]{\todo[linecolor=blue,backgroundcolor=blue!25,bordercolor=blue,#1]{#2}}
\newcommandx{\info}[2][1=]{\todo[
linecolor=green,backgroundcolor=green!25,bordercolor=green,#1]{#2}}

\usepackage{booktabs} 

\setcopyright{rightsretained}

\acmDOI{10.1145/nnnnnnn.nnnnnnn}

\acmISBN{978-x-xxxx-xxxx-x/YY/MM}

\acmConference[GECCO '20]{the Genetic and Evolutionary Computation Conference 2020}{July 8--12, 2020}{Canc\'un, Mexico}
\acmYear{2020}
\copyrightyear{2020}

\acmPrice{15.00}

\copyrightyear{2020}
\acmYear{2020}
\setcopyright{rightsretained}
\acmConference[GECCO '20]{Genetic and Evolutionary Computation Conference}{July 8--12, 2020}{Canc\'un, Mexico}
\acmBooktitle{Genetic and Evolutionary Computation Conference (GECCO '20), July 8--12, 2020, Canc\'un, Mexico}\acmDOI{10.1145/3377930.3390221}
\acmISBN{978-1-4503-7128-5/20/07}

\begin{document}

\title{Discovering Representations for Black-box Optimization} 

\author{Adam Gaier}
\affiliation{%
  \institution{Inria, CNRS, Universit\'e de Lorraine}  
  \institution{Bonn-Rhein-Sieg University of Applied Sciences}  
}
\email{adam.gaier@h-brs.de}

\author{Alexander Asteroth}
\affiliation{%
  \institution{Bonn-Rhein-Sieg University of Applied Sciences}
  \city{Sankt Augustin} 
  \country{Germany}
  \postcode{53757}
}
\email{alexander.asteroth@h-brs.de}

\author{Jean-Baptiste Mouret}
\affiliation{%
  \institution{Inria, CNRS,}
  \institution{Universit\'e de Lorraine}
  \city{Nancy} 
  \country{France}\postcode{54000}}
\email{jean-baptiste.mouret@inria.fr}

\begin{abstract}

The encoding of solutions in black-box optimization is a delicate, handcrafted balance between expressiveness and domain knowledge --- between exploring a wide variety of solutions, and ensuring that those solutions are useful.
Our main insight is that this process can be automated by generating a dataset of high-performing solutions with a quality diversity algorithm (here, MAP-Elites), then learning a representation with a generative model (here, a Variational Autoencoder) from that dataset. 
Our second insight is that this representation can be used to scale quality diversity optimization to higher dimensions --- but only if we carefully mix solutions generated with the learned representation and those generated with traditional variation operators. 
%
%
We demonstrate these capabilities by learning an low-dimensional encoding for the inverse kinematics of a thousand joint planar arm.
The results show that learned representations make it possible to solve high-dimensional problems with orders of magnitude fewer evaluations than the standard MAP-Elites, and that, once solved, the produced encoding can be used for rapid optimization of novel, but similar, tasks.
The presented techniques not only scale up quality diversity algorithms to high dimensions, but show that black-box optimization encodings can be automatically learned, rather than hand designed.

\end{abstract}

\maketitle



\begin{figure}[t]
  \centering
  \includegraphics{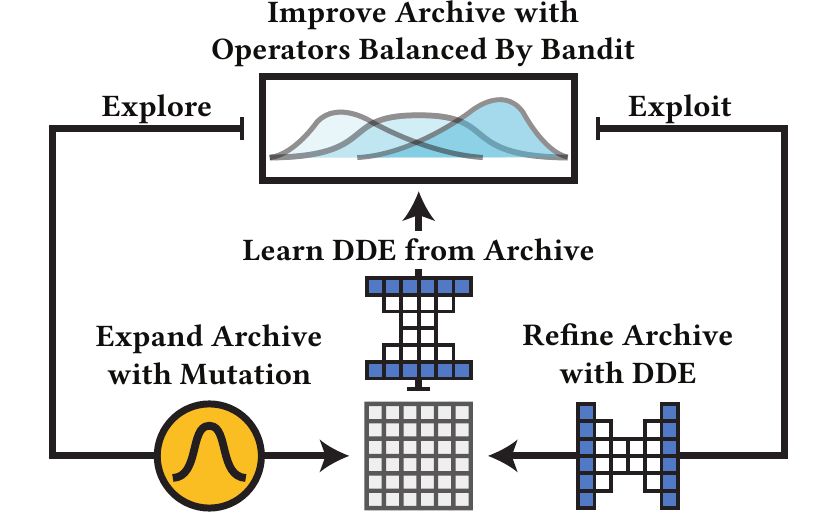}
  \caption
  { 
    \textit{Data-Driven Encoding MAP-Elites} (DDE-Elites) searches the space of representations to search for solutions. A data-driven encoding (DDE) is learned by training a VAE on the MAP-Elites archive. 
    High fitness solutions, which increase the bias of the DDE toward performance, are found using the DDE. Novel solutions, which increase the range of solutions which can be expressed, are found using mutation operators. UCB1, a bandit algorithm, balances the mix of these explorative and exploitative operators.
    }
  \label{fig:overview}
\end{figure} 
\section{Introduction}


%
The method of encoding solutions is one of the most critical design decisions in optimization, as the representation defines the way an algorithm can move in the search space~\cite{rothlauf2006representations}. 
Work on representations tends to focus on encoding priors or innate biases: 
aerodynamic designs evolved with splines to encourage smooth forms \cite{olhofer2001adaptive}, 
Compositional Pattern Producing Networks (CPPNs) with biases for symmetry and repetition in images and neural network weight patterns \cite{cppn,hyperneat}, 
modularity induced in evolved neural networks~\cite{mouret2008mennag,durr2010genetic,doncieux2004evolving}, or 
neural network structures which encode strong enough biases to perform without training~\cite{gaier2019weight}.  

The best representations balance a bias for high performing solutions, so they can easily be discovered, and the ability to express a diversity of potential solutions, so the search space can be widely explored.
At the one extreme, a representation which only encodes the global optimum is easy to search but useless for finding any other solution. At the other, a representation which can encode anything presents a difficult and dauntingly vast search space.

Given a large set of example solutions, representations could be learned from data instead of being hand-tailored by trial-and-error: a learned representation would replicate the same biases toward performance and the same range of expressivity as the source data set. For instance, given a dataset of face images, a Variational Autoencoder (VAE) \cite{vae} or a Generative Adversarial Network (GAN) \cite{gan} can learn a low-dimensional latent space, or encoding, that makes it possible to explore the space of face images. In essence, the decoder which maps the latent space to the phenotypic space learns the ``recipe'' of faces. Importantly, the existence of such a low-dimensional latent space is possible because \emph{the dataset is a very small part of the set of all possible images}.

However, using a dataset of preselected high-performing solutions ``traps'' the search within the distribution of solutions that are already known: a VAE trained on white faces will never generate a black face. This limits the usefulness of such data-driven representations for discovering \emph{novel} solutions to hard problems.

In this paper, we propose the use of the MAP-Elites algorithm \cite{mapelites} to automatically generate a dataset for representations using only a performance function and a diversity space. 
Quality diversity (QD) algorithms~\cite{qd2,qd} like MAP-Elites are a good fit for representation discovery: creating archives of diverse high-performing solutions is precisely their purpose.
Using the MAP-Elites archive as a source of example solutions, we can capture the genetic distribution of the highest performing solutions, or elites, by training a VAE and obtaining a latent representation. 
As the VAE is only trained on elites, this learned representation, or Data-Driven Encoding (DDE), has a strong bias towards solutions with high fitness; and because the elites have varying phenotypes, the DDE is able to express a range of solutions. Though the elites vary along a phenotypic continuum, they commonly have many genotypic similarities~\cite{me_linemut}, making it more likely to find a well-structured latent space.

Nonetheless, MAP-Elites will struggle to find high-performing solutions without an adequate representation. Fortunately, the archive is produced by MAP-Elites in an iterative, any-time fashion, so there is no ``end state'' to wait for before a DDE can be trained --- a DDE can be trained \textit{during optimization}. The DDE can then be used to enhance optimization. By improving the quality of the archive the DDE improves the quality of its own source data, establishing a virtuous cycle of archive and encoding improvement.

A DDE based on an archive will encounter the same difficulty as any learned encoding:
the DDE can only represent solutions that are already in the dataset. How then, can we discover new solutions? Fundamentally, to search for an encoding we need to both \emph{exploit the best known representation}, that is, create better solutions according to the current best ``recipes'', and also \emph{explore new representations} --- solutions which do not follow any ``recipe''. 
%

In this paper, we address this challenge by mixing solutions generated with the DDE with solutions obtained using standard evolutionary operators. Our algorithm applies classic operators, such as Gaussian mutation, to create candidates which could not be captured by the current DDE. At the same time we leverage the DDE to generalize common patterns across the map and create new solutions that are likely to be high-performing. To avoid introducing new hyper-parameters, we tune this exploration/exploitation trade-off optimally using a multi-armed bandit algorithm \cite{garivier2011upper}.

This new algorithm, DDE-Elites, reframes optimization as a search for representations (Figure \ref{fig:overview}). Integrating MAP-Elites with a VAE makes it possible to apply  quality diversity to high-dimensional search spaces, and to find effective representations for future uses.
%
We envision application to domains that have straightforward but expansive low-level representations, for instance: joints positions at 20Hz for a walking robot ($12 \times 100=1200$ joint positions for a 5-second gait of a robot with $12$ degrees of freedom), 3D shapes in which each voxel is encoded individually (1000-dimensional for a $10 \times 10 \times 10$ grid), images encoded in the pixel-space, etc.

Ideally, the generated DDE will capture the main regularities of the domain. In robot locomotion, this could correspond to periodic functions, since we already know that a $36$-dimensional controller based on periodic functions can produce the numerous joint commands required every second to effectively drive a 12-joint walking robot in many different ways~\cite{cully2015robots}. In many domains the space of possible solutions can be vast, while the inherent dimensionality of interesting solutions is still compact. By purposefully seeking out a space of solutions, rather than the solutions themselves, we can solve high-dimensional problems in a lower dimensional space.


\section{Background}
\subsection{Optimization of Representations} 
  In his 30 year perspective on adaptation in evolutionary algorithms, Kenneth De Jong identified representation adaptation as "perhaps the most difficult and least understood area of EA design."~\cite{de2007parameter}

  Despite the difficulty of creating adaptive encodings, the potential rewards have lured researchers for decades. Directly evolving genotypes to increase in complexity has a tradition going back to the eighties~\cite{goldberg1989messy,altenberg1994evolving}. The strategy of optimizing a solution at low complexity and then adding degrees of freedom has proved effective on problems from optimal control~\cite{gaier2014evolution}, to aerodynamic design~\cite{olhofer2001adaptive}, to neural networks~\cite{neat}. 
  Evolving the genome's structure is particularly important when the structure itself is the solution, such as in genetic programming~\cite{koza1990genetic} or neural architecture search~\cite{elsken2019neural,miikkulainen2019evolving,gaier2019weight}.

  Recent approaches toward representation evolution have focused on genotype-phenotype mappings~\cite{bongard2003evolving}. Neural networks, which map between inputs and outputs, are a natural choice for such `meta-representations'. These mappings can evolve with the genome~\cite{scott2015learning,simoes2014self}, or fix the genome and evolve only the mapping~\cite{hyperneat,cppn}. 

  Supervised methods have been previously applied to learn encodings. These approaches require a set of example solutions for training. Where large, well-curated data sets are available this strategy has proven effective at creating representations well suited to optimization~\cite{mariogan,latentVariableEvolution2_fingerprint,latentVariableEvolution1_art}, but where a corpus of solutions does not exist it must be created. In \cite{scott2018toward,moreno2018learning} these solutions were collected by saving the champion solutions found after repeatedly running an optimizer on the problem, with the hope that the learned representation would then be effective in similar classes of problems.
\subsection{MAP-Elites}

  MAP-Elites~\cite{mapelites} is a QD algorithm which uses a niching approach to produce high-performing solutions which span a continuum of user-defined phenotypic dimensions. These phenotypic dimensions, or behavior descriptors, describe \textit{the way} the problem is solved, and are often orthogonal to performance.
  MAP-Elites has been used in such diverse cases as optimizing the
  distance traveled by a walking robot using different legs~\cite{cully2015robots},
  the drag of aerodynamic designs with varied volumes and curvatures~\cite{gaier2017aerodynamic},
  and the win rate of decks composed of different cards in deck-building games~\cite{fontaine2019mapping}. 

  MAP-Elites is a steady-state evolutionary algorithm which maintains a population in a discretized grid or `archive'. This grid divides the continuous space of possible behaviors into bins, or `niches' with each bin holding a single individual, or `elite'. These elites act as parents, and are mutated to form new individuals. These child individuals are evaluated and assigned a niche based on their behavior. If the niche is empty the child is placed inside; if the niche is already occupied, the individual with higher fitness is stored in the niche and the other discarded. By repeating this process, increasingly optimal solutions which cover the range of phenotype space are found. The MAP-Elites algorithm is summarized in Algorithm~\ref{alg:mapelites}.

  \begin{algorithm} [h!]
    \caption{MAP-Elites}
    \label{alg:mapelites}
    \begin{algorithmic}[1]
      \Function{MAP-Elites}{$fitness()$, $variation()$, $\mathcal{X}_{initial}$}
      \State $\mathcal{X}\gets\emptyset$, $\mathcal{F}\gets\emptyset$ 
      \Comment{\textit{Map of genomes $\mathcal{X}$, and fitnesses $\mathcal{F}$}}
     
      \State $\mathcal{X} \gets \mathcal{X}_{initial}$
      \Comment{\textit{Place initial solutions in map}}
      \State $\mathcal{F} \gets fitness(\mathcal{X}_{initial})$

        \For{iter = $1 \to I$}
          \State $\mathbf{x'} \gets variation(\mathcal{X})$ \Comment{\textit{Create new solution from elites}}
          \State $\mathbf{p'}, \mathbf{b'}  \gets fitness(\mathbf{x'})$
          \Comment{\textit{Get performance and behavior}}
          \If{$\mathcal{F}(\mathbf{b'}) = \emptyset$ or $\mathcal{F}(\mathbf{b'}) < \mathbf{f'}$}
          \Comment{\textit{Replace if better}}        
            \State $\mathcal{F}(\mathbf{b'}) \gets \mathbf{f'}$
            \State $\mathcal{X}(\mathbf{b'}) \gets \mathbf{x'}$
          \EndIf
        \EndFor
        \State \Return $(\mathcal{X}$, $\mathcal{F})$ 
        \Comment{\textit{Return illuminated map}}
        \EndFunction
    \end{algorithmic}
  \end{algorithm}

  Though phenotypically diverse the elites are often genotypically similar, existing in an ``elite hypervolume'', a high performing region of genotype space~\cite{me_linemut}. Just as in nature, where species as diverse as fruit flies and humans share nearly 60 percent of their genome~\cite{fruitfly}, the ``recipe'' for high performance is often composed of many of the same ingredients.


  This insight was leveraged in ~\cite{me_linemut} to create a new variation operator which considers the correlation among elites. Genes which vary little across the elites, and so are likely common factors that produce high performance, are also subject to the smallest amount of perturbation --- lowering the chance their children stray from the elite hypervolume. Biasing mutation in this way ensures that exploration is focused on factors which induce phenotypic variation without drifting into regions of poor performance.

\subsection{Variational Autoencoders}

  Autoencoders (AEs)~\cite{ae} are neural networks designed to perform dimensionality reduction. AEs are composed of two components: an encoder, which maps the input to a lower dimensional latent space; and a decoder, which maps the latent space back to the original space. The decoder is trained to reconstruct the input through this lower dimensional latent ``bottleneck''. The encoder component can be viewed as a generalization of Principal Component Analysis~\cite{pca}, with the latent space approximating principal components.

  Though the AE is able to represent the data at a lower dimensionality, and reproduce it with minimal loss, it can still be a poor representation for optimization. An important quality of representations is `locality', that a small change in the genotype induces a small change in the phenotype~\cite{rothlauf2006representations}. When AEs are trained only to minimize reconstruction error they may overfit the distribution of the training data and create an irregular latent space. The low-locality of such latent spaces limits their usefulnesses in optimization: nearby points in latent space may decode to very different solutions, meaning even a small mutation could have a large effect.

  Variational autoencoders (VAEs)~\cite{vae} are AEs whose training is regularized to ensure a high-locality latent space. The architecture is broadly the same: an encoder and decoder mediated by a bottleneck, but rather than encoding the input as a single point it is encoded as a normal distribution in the latent space. When training the model a point from this input distribution is sampled, decoded, and the reconstruction error computed. By encoding the input as a normal distribution we induce the distributions produced by the encoder to be closer to normal. VAEs are trained by minimizing two terms: (1) the reconstruction error, and (2) the Kullback-Liebler (KL) divergence~\cite{kldivergance}~  
  of the latent space to a unit Gaussian distribution, giving the loss function:
  
  \begin{equation}
  loss =\|x-\hat{x}\|^{2}+K L\left[N\left(\mu_{x}, \sigma\right), N(0,1)\right]   
  \end{equation}

  Inducing solutions to be encoded in the form of a normal distribution structures the latent space in a continuous and overlapping way, creating a local encoding better suited to optimization.  

\section{DDE-Elites}

\begin{figure*}[ht!]
  \centering
  \includegraphics{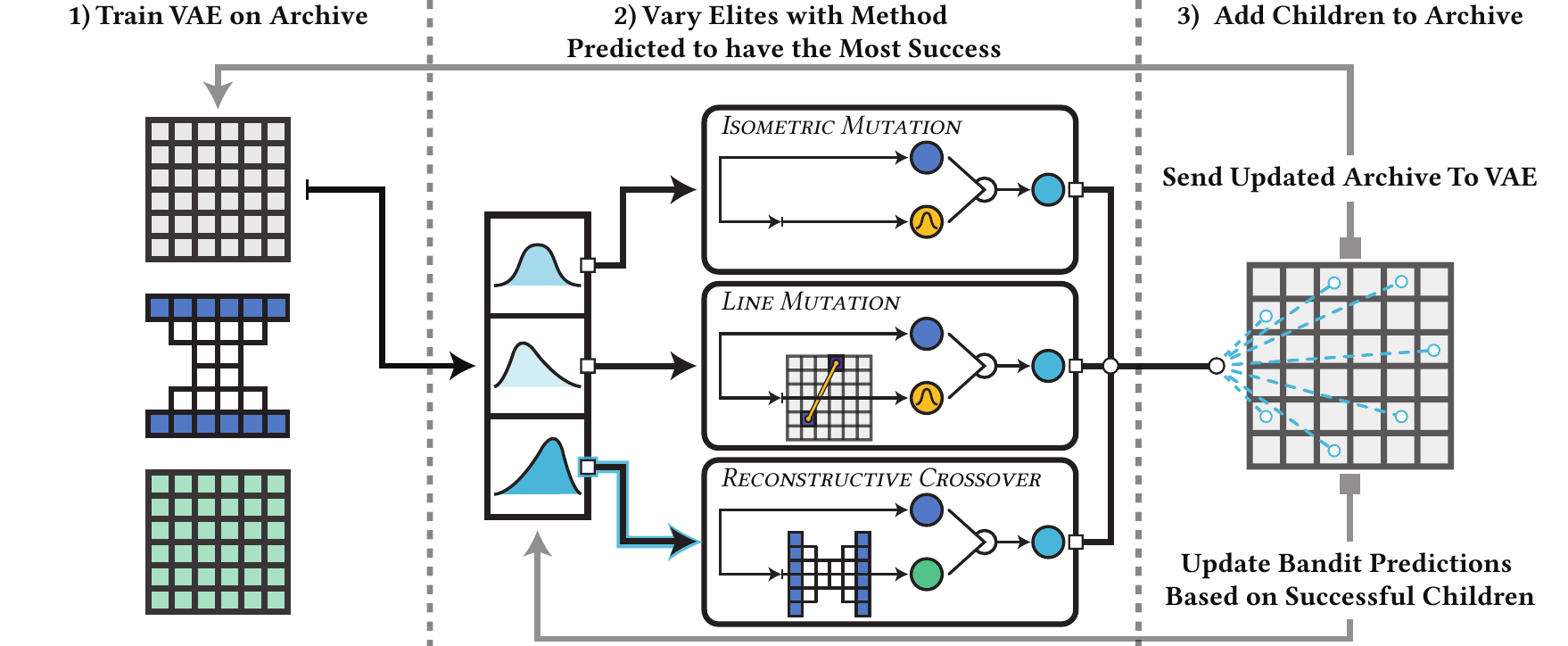}
  \caption
  { 
    \textit{DDE-Elites Algorithm}
    \newline
    (1) A VAE is trained on the archive, and used to create a `reconstructive crossover' operator which creates new solutions by averaging the parameters of an individual with its own reconstruction; (2) the mix of exploitative and explorative variation operators predicted to have the most success is chosen by the multi-armed bandit algorithm UCB1 and used to create new solutions; (3) the new solutions are added to the archive  and the success rate of the applied variation operator is updated.
    }
  \label{fig:explainer}
\end{figure*}

    Every representation biases optimization in some way,
    improving optimization by limiting the range of solutions that can be expressed to those which are valid or high-performing~\cite{rothlauf2006representations}. 
    But finding a balance between expressivity and bias is an arduous task requiring considerable domain expertise.
    Our method, DDE-Elites, automates the process of representation design and learns new encodings in tandem with search --- allowing optimization and representation learning to improve each other in a self-reinforcing cycle.

    DDE-Elites learns an encoding from examples of high performing solutions.
    To create these examples we use MAP-Elites, which produces a variety of high performing solutions rather than converging to a single optima. 
    ~The variety produced by MAP-Elites is critical --- the expressivity of any learned encoding is limited by the variety of examples. That MAP-Elites not only produces a variety of solutions, but allows us to define the nature of that variety, makes it particularly powerful for crafting useful representations. By defining the type of variety we want to explore we are defining the biases and expressivity we encode in our representation.
  
    DDE-Elites is a variant of the MAP-Elites algorithm. The core component of competition within a niched archive is maintained, but novel methods of producing child solutions are introduced. Child solutions are created using an encoding learned from the archive. This encoding is refined as the archive improves, which in turn improves the optimization process. DDE-Elites optimizes an archive of varied solutions by reframing optimization as a search for the best representation, rather than the best solution.

    The DDE-Elites algorithm proceeds as follows (see Figure~\ref{fig:explainer} and Algorithm~\ref{alg:dde-elites}): (1) a DDE and \textit{reconstructive crossover} operator is created by training a VAE on the archive; (2) the probability of using each variation operator is determined by the UCB1 bandit algorithm; (3) MAP-Elites is run with the chosen variation operator probabilities. The success rate of the variation operators to create solutions is used to update the bandit and the improved archive is used to create a new DDE and reconstructive crossover operator.

  \paragraph{Data Driven Encoding}
  The MAP-Elites archive is a record of the highest-performing solutions yet found in each bin.
  When the archive is updated the VAE is trained to reconstruct the individuals in the archive. 
  Reconstruction is a mapping from one phenotype to another, mediated through latent space; and the mapping from latent space to phenotype space analogous to a genotype-phenotype mapping, which we refer to as a Data-Driven Encoding (DDE).

  Features common in high performing solutions will be the most successfully compressed and reconstructed --- and features widely shared by high performing solutions are likely to lead to high performance.
  Critically, by training the encoding only on high-performing solutions we bias the space of solutions the DDE can express to those with high performance.

  \paragraph{Reconstructive Crossover}
  By limiting the range of solutions which can be expressed by a representation, we are able to bias the solutions found during search.
  When a solution is reconstructed with the VAE it is mapped onto the restricted space of solutions expressible by the DDE --- a space characterized by high performance.

  Reconstructing individuals with the VAE can create new solutions with higher fitness than the originals, but cannot create novel solutions.
  Solutions created by the DDE are based on those already in the archive, so cannot reach solutions which lie outside of the encoded distribution.
  At early stages of optimization when there are few example solutions, using only reconstruction to create new solutions would doom our encoding to a small region of expression.

  Rather than completely replacing individuals with their reconstructions we instead shift them closer to forms expressible by the DDE with a new variation operator, \textit{reconstructive crossover}. Child solutions are created by performing crossover with two parents: a parent chosen from the archive and its reconstruction. Crossover takes the form of an element-wise mean of the parameter vectors. 

  \begin{equation}
    \mathbf{x}_{i}^{(t+1)}= \frac{1}{2}*(\mathbf{x}_{i}^{(t)} + 
                    VAE.Decode(VAE.Encode(\mathbf{x}_{i}^{(t)})) )
  \end{equation}

  The reconstructive crossover operator slows the loss of diversity by only moving an individual \textit{toward} the distribution of solutions encoded by the DDE, not directly into it. By only shifting solutions rather than replacing them, we allow exploration outside of the distribution to continue.
  ~%
  Even when there is little gain in fitness, solutions that are the result of reconstructive crossover have a lower inherent dimensionality, on the account of having parents pass through the compressive bottleneck of the VAE. In this way the reconstructive crossover operator not only spreads globally advantageous genes throughout the archive, but also pulls the archive towards more easily compressed solutions.

  \paragraph{Line Mutation}
  Reconstructive crossover enables effective optimization within the range of solutions that the DDE can express, but explorative operators are required to widen the pool of example solutions and improve the DDE. So when creating new solutions we choose to either produce them through reconstructive crossover, or through random mutation.


  In addition to isometric Gaussian mutation commonly used in MAP-Elites, we apply the line mutation operator proposed in~\cite{me_linemut}. Line mutation imposes a directional component on the Gaussian perturbations.
  During mutation the parent genome is compared to a random genome from the archive.
  The variance of mutation in each dimension is then scaled by the difference in each gene: 
  \begin{equation}
    \mathbf{x}_{i}^{(t+1)}=\mathbf{x}_{i}^{(t)}+\sigma_{1} \mathcal{N}(0, \mathbf{I})+\sigma_{2}\left(\mathbf{x}_{j}^{(t)}-\mathbf{x}_{i}^{(t)}\right) \mathcal{N}(0,1)
  \end{equation}
  where $\sigma_{1}$ and $\sigma_{2}$ are hyperparameters which define the relative strength of the isometric and directional mutations. Intuitively, when two genes have similar values the spread of mutation will be small, when the values are very different the spread will be large.

  In many cases certain parameter values will be correlated to high fitness, regardless of the individual's place in behavior space. The line operator is a simple way of exploiting this similarity, but in contrast to reconstructive crossover does not limit expressivity -- allowing it to be used as a method of exploring new solutions. Though both the reconstructive crossover and line mutation operators take advantage of the similarities between high performing individuals, their differing approaches allow them to be effectively combined as explorative and exploitative operators.

  \paragraph{Parameter Control}
  DDE-Elites explores the space of representations with the exploitative operator of reconstructive crossover, which finds high performing solutions similar to those already encoded by the DDE, and explorative operators of mutation, which expand the space of solutions beyond the range of the DDE.

  The optimal ratio to use these operators is not only domain dependent, but dependent on the stage of the algorithm. When the archive is nearly empty, it makes little sense to base a representation on a few randomly initialized solutions; once the behavior space has been explored, it is beneficial to continue optimization through the lens of the DDE; and when the archive is full of solutions produced by the DDE it is more useful to expand the range of possible solutions with mutation.  These stages are neither predictable nor clear cut, complicating the decision of when to use each operator. 

  Faced with a trade-off between exploration and exploitation we frame the choice of operators as a multi-armed bandit problem~\cite{auer2002finite}. Multi-armed bandits imagine sets of actions as levers on a slot machine, each with their own probability of reward. The goal of a bandit algorithm is to balance exploration, trying new actions, and exploitation, repeating actions that yield good rewards. Bandit approaches are straightforward to implement and have been previously used successfully to select genetic operators~\cite{dacosta2008adaptive}.

  We define a set of possible actions as usage ratios between reconstructive crossover, line mutation, and isometric mutation. The ratio of $[\frac{1}{4},\frac{3}{4},0]$, for example, would have solutions created by reconstructive crossover with a probability of $\frac{1}{4}$, line mutation with a probability of $\frac{3}{4}$, and never with isometric mutation. Each action is used to create a batch of child solutions and a reward is assigned in proportion to the number of children who earned a place in the archive. At each generation a new action is chosen, and the reward earned for that action recorded.

  Actions are chosen based on UCB1~\cite{auer2002finite}, a simple and effective bandit algorithm which minimizes regret. Actions with the greatest potential reward are chosen, calculated as:
  \begin{equation}
  Q(a)+\sqrt{(2 \log t) / (N_{t}(a))}
  \end{equation}
  where $Q(a)$ is the reward for an action $a$, $t$ is the total number of actions that have been performed, and $N_{t}(a)$ the number of times that action has been performed. UCB1 is an optimistic algorithm which rewards uncertainty --- 
  given two actions with the same mean reward, the action which has been tried fewer times will be chosen. 


  Our archive is in constant flux, and so the true reward of each mix of operators changes from generation to generation. To handle the non-stationary nature of the problem we use a sliding window~\cite{garivier2011upper}, basing our predictions only on the most recent generations.


  \begin{algorithm} [h]
    \caption{DDE-Elites}
    \label{alg:dde-elites}
    \begin{algorithmic}[1]
    \Function{DDE-Elites}{$fitness()$ $\mathcal{X}_{initial}$}
    \State $\mathcal{X} \gets \mathcal{X}_{initial}$
    \State $\mathcal{V}$: Possible Variation Operator Probabilities (vector)
    \State (e.g., [0,0.5,0.5], [0.8,0.0,0.2], [1.0,0.0,0.0] for [xover,line,iso])
    \State successes $\gets zeros(len(\mathcal{V}))$ \Comment{\# successes for each option}
    \State selection $\gets zeros(len(\mathcal{V}))$ \Comment{\# selections for each option}
    
    \For{iter = $1 \to I$}
      \State \HiLi \textit{ --- Train VAE on Current Archive --- }
      \State VAE.Train~($\mathcal{X}$)

      \State \HiLi \textit{ --- Choose Variation Based on UCB1 --- }
      \State $i \leftarrow \arg \max\left(\frac{\text {successes}[s]}{\text { selected }[s]}+\sqrt{\frac{2 \ln (\text{sum}(\text{successes}))}{\text {selected}[s]}}\right)$

      \State \HiLi \textit{ --- Run MAP-Elites Using Chosen Variation --- }
      \State $variation() \gets \mathcal{V}[i]$

      \State $\mathcal{X'} \gets $MAP-Elites$(fitness(), variation(), \mathcal{X}$)

      \State \HiLi \textit{ --- Track Performance of Chosen Variation --- }
      \State $selection[i] \gets selection[i] + 1$
      \State $successes[i] \gets successes[i] + nImproved(\mathcal{X'}, \mathcal{X})$
    \EndFor
    \State DDE $\gets$ VAE.Decode()
    \State \Return $\mathcal{~X}$, $\text{DDE}$ 

    \EndFunction
    \end{algorithmic}
    \vspace{0.0cm}

    \begin{algorithmic}[1]
    \Function{Isometric Mutation}{$\mathcal{X}$} 
      \State $\mathbf{x~} \gets random\_selection(\mathcal{X})$
      \State \Return $\mathbf{x} + \sigma     \mathcal{N}(0, \mathbf{I})$
    \EndFunction
    \end{algorithmic}

    \vspace{0.0cm}
    \begin{algorithmic}[1]
    \Function{Line Mutation}{$\mathcal{X}$} 
      \State $\mathbf{x~,y~} \gets random\_selection(\mathcal{X})$
      \State \Return $\mathbf{x}+ \sigma_{1} \mathcal{N}(0, \mathbf{I}) 
                                + \sigma_{2} (\mathbf{x}-\mathbf{y}) \mathcal{N}(0,1)$
    \EndFunction
    \end{algorithmic}

    \vspace{0.0cm}
    \begin{algorithmic}[1]
    \Function{Reconstructive Crossover}{$\mathcal{X}$} 
      \State $\mathbf{x~} \gets random\_selection(\mathcal{X})$
      \State $\mathbf{y~} \gets VAE.Decode(VAE.Encode(\mathbf{x}))$ \Comment{VAE Reconstruction}
      \State \Return $(\mathbf{x} + \mathbf{y})/2$
    \EndFunction
    \end{algorithmic}
  \end{algorithm}


\begin{figure*}[t!]
  \centering
  \includegraphics{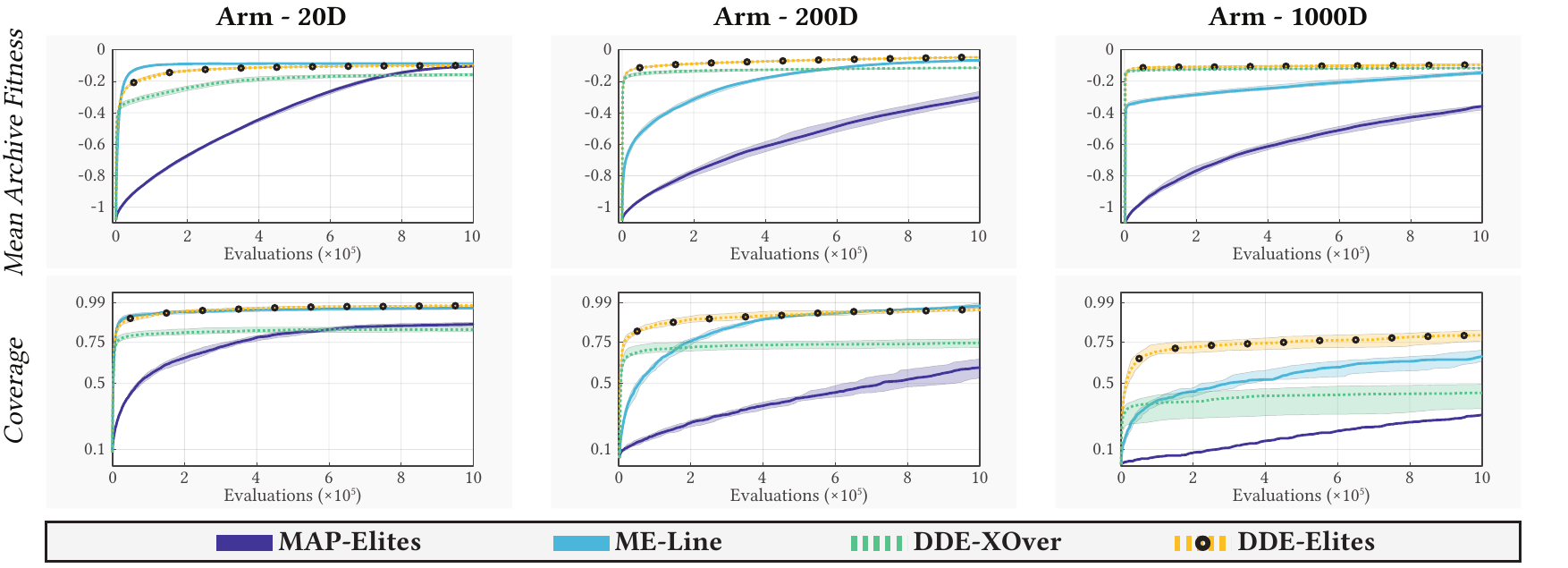}
  \caption
  { 
    \textit{Archive Illumination}
    \newline
    Archive illumination performance of MAP-Elites with different variation operators: standard isometric mutation (\textit{MAP-Elites}), line mutation (\textit{ME-Line}), reconstructive crossover (\textit{DDE-XOver}) and DDE-Elites, which uses the UCB1 bandit algorithm to choose between the three at every generation. We measure fitness as the mean fitness of all solutions in the archive; coverage as the fraction of behavior space bins which contain solutions. Results over 20 replicates with lines indicating medians and quartile bounds shaded. The median of DDE-Elites, our approach, is additionally noted with black dots. All final results are significantly different ($p < 0.01$ Mann-Whitney U) in fitness and coverage. Progress is shown in evaluations (0 to 1 million); a batch size of 100 evaluations per generation was used, so this scale corresponds to generations from 0 to 10,000.
  }
  \label{fig:archive}
\end{figure*}

\section{Experiments}
\begin{figure*}[t!]
  \centering
  \includegraphics{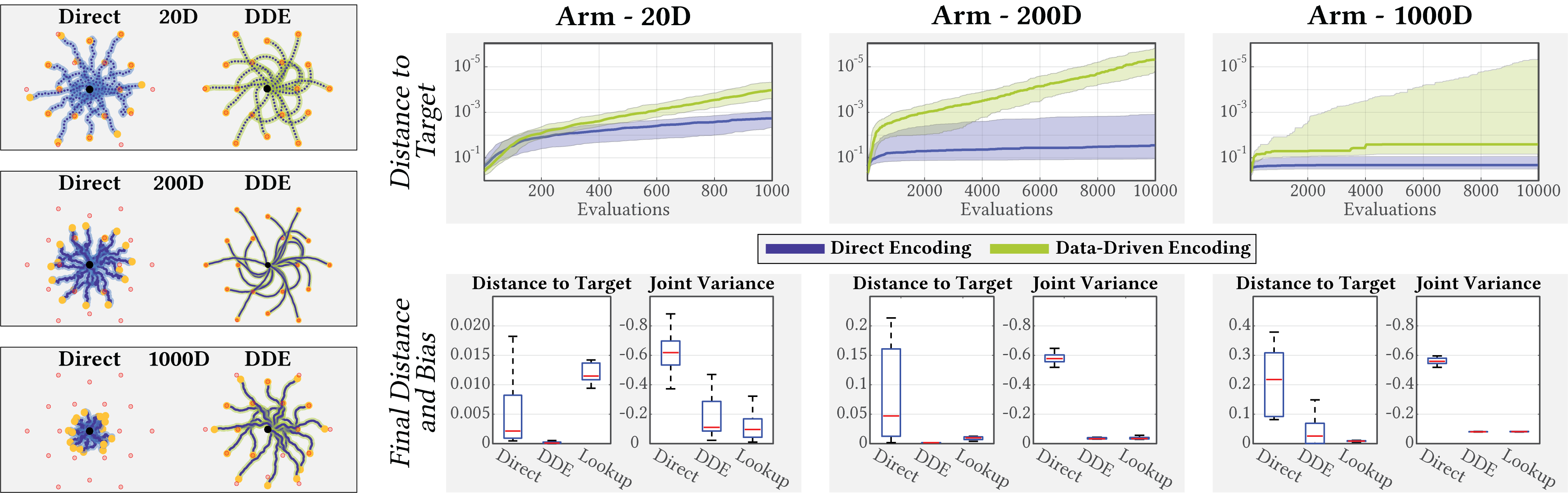} 
  \caption
  { 
    \textit{Optimization with Direct and Data-Driven Encodings}
    \newline
    CMA-ES is given a set budget to find a solution with a target behavior, and searches with either a direct encoding or a DDE. 
    \\ \textit{Left:} Example solutions for target matching with the direct and data driven encodings. End effectors in yellow, targets in red.
     \\ \textit{Top:} Optimization over time of median distance (dotted line) to the 18 targets over 50 replicates (quartiles shaded).
     \\ \textit{Bottom:} The final distance to the targets, and a characteristic of the solution. These characteristics were not optimized by CMA-ES, but optimized during the creation of the DDE, biasing the solutions produced.
    }
  \label{fig:targetMatch}
\end{figure*}

\paragraph{Planar Arm Inverse Kinematics}\footnote{see Figure ~\ref{fig:targetMatch} for a visualization of this domain}
    We demonstrate the effectiveness of DDEs and DDE-Elites on in the inverse kinematics (IK) problem of a 2D robot arm, a common QD benchmark problem~\cite{qd2,me_linemut}. Given target coordinates a configuration of joint angles should be found to place the end effector at the target. To solve this task, a discretized behavior space is defined over the x,y plane and MAP-Elites finds a configuration of joint angles which places the end effector in each bin. The location of the end effector is derived for an arm with $n$ joints with angles $y$ with using the forward kinematics equation:
    \vspace{-0.35cm}

    \begin{equation}
      \mathbf{b}(\mathbf{y})=\left[\begin{array}{c}
      {l_{1} \cos(y_{1})+l_{2}\cos(y_{1}+y_{2})+\cdots+l_{n}\cos(y_{1}+\cdots+y_{n})} \\
      {l_{1} \sin(y_{1})+l_{2}\sin(y_{1}+y_{2})+\cdots+l_{n}\sin(y_{1}+\cdots+y_{n})}
      \end{array}\right]\nonumber
    \end{equation}

    There are many solutions to this IK problem, but solutions with lower joint variance are preferred to allow for smoother transitions between configurations. We define fitness as the negative joint variance:~$-\frac{1}{n} \sum_{i=1}^{n} (y_{i}-\mu)^2$ where $(\mu=\sum_{i=1}^{n} y_{i})$.

    To summarize: the phenotype is the angle of each joint, the behavior is the x,y coordinates of the end effector, and the fitness the negative variance of the joint angles. The difficulty of the problem can be easily scaled up by increasing the number of joints in the arm: we solve this task with 20, 200, and 1000 joints. When a DDE is used 10 latent dimensions are used for the 20D arm, and 32 dimensions for the 200 and 1000D arms. The same archive structure is used for all domains. A unit circle is divided into 1950 bins, with each bin defined by the Voronoi cell~\cite{cvt} with centers placed in a ring formation\footnote{See supplementary material for a visualization of this structure}.


\begin{figure*}[t!]
  \centering
  \includegraphics{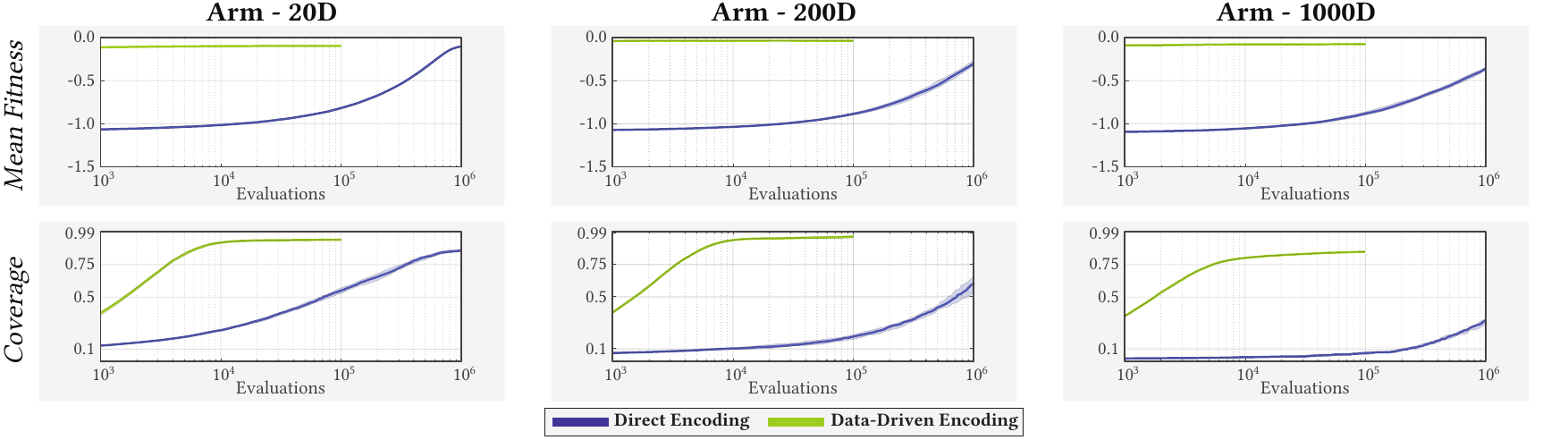} 
  \caption
  { 
    \textit{Archive Recreation with Data-Driven Encoding}
    \newline
    Performance of MAP-Elites algorithm when run with direct or data-driven encoding. When using the direct encoding, MAP-Elites was given one order of magnitude more evaluations (note logarithmic scale of evaluations). Fitness is measured as the mean fitness of all solutions in the archive, coverage as the fraction of behavior space bins which contain solutions. Results over 50 replicates with dotted lines indicating medians and quartile bounds shaded.
    }
  \label{fig:recreate}
\end{figure*}

\subsection{Archive Illumination}
    We first demonstrate the ability of DDE-Elites to scale up illumination to high-dimensional problems. The performance of DDE-Elites is compared to three algorithmic variants: the canonical MAP-Elites algorithm using isometric mutation (\textit{MAP-Elites}); MAP-Elites using line, or directional, mutation (\textit{ME-Line}); and MAP-Elites using the reconstructive crossover (\textit{DDE-XOver}).  Our proposed approach \textit{DDE-Elites} uses all operators at a ratio determined by the UCB1 bandit algorithm. These treatments are summarized in Table~\ref{tab:algVariants}.

    \begin{table}[h!]
    \begin{tabular}{|r|c|c|c|}
    \hline
    \multicolumn{1}{|l|}{}  & 
    \textit{\begin{tabular} [c]{@{}c@{}}Isometric\\ Mutation\end{tabular}} &
    \textit{\begin{tabular}[c]{@{}c@{}}Line \\ Mutation\end{tabular}} & 
    \textit{\begin{tabular}[c]{@{}c@{}}Reconstructive\\ Crossover\end{tabular}} \\ \hline
    \textbf{MAP-Elites} & X &  &  \\ \hline
    \textbf{ME-Line} &  & X &  \\ \hline
    \textbf{DDE-XOver} &  &  & X \\ \hline
    \textit{\textbf{DDE-Elites}} & X & X & X \\ \hline
    \end{tabular}
    \caption{Algorithm variants. DDE-Elites is our approach.}
    \label{tab:algVariants}
    \end{table}

    These variants are compared based on the quality of the archive at each generation~(Figure \ref{fig:archive}). Archives are judged based on two metrics: (1) coverage, the number of bins filled, and (2) performance, the mean fitness of solutions.\footnote{Sixty-four core machines were used to evaluate 100 individuals in parallel, requiring $\sim$0.2s, $\sim$0.8s, $\sim$1.6s, for the arm at 20d, 200d, and 1000D arm respectively. In every case the VAE required $\sim$2.4s to train on a single CPU core.}

    In the 20-dimensional case ME-Line quickly fills the map with high performing solutions. In only a one hundred thousand evaluations ME-Line creates an archive unmatched by MAP-Elites even after one million evaluations. When only the reconstructive crossover operator is used, despite promising early progress, a chronic lack of exploration results in archives which are worse than the standard MAP-Elites. DDE-Elites, with access to all operators, explores as quickly as ME-Line and creates archives of similar quality.

    When the dimensionality of the arm is scaled up to 200D, we see the convergence rate of ME-Line slow down considerably. While still reaching high levels of performance it does so only after one million evaluations, a tenth of the evaluations required in in the 20D case --- suggesting that the effectiveness of ME-Line scales linearly with the dimensionality of the problem. In contrast DDE-Elites is barely affected by a ten-fold increase in parameters --- exploration is only slightly slowed, and high-performing solutions are found from the very earliest iterations. 
    %
    The effects of scaling can be observed even more clearly in the 1000D case: ME-Line illuminates the archive only very slowly, while the performance of DDE-Elites is marked by the same burst of exploration and consistently high fitness solutions that characterized its performance in lower dimensions.


    The line mutation operator is clearly able to leverage the similarities in high performing solutions across the archive --- in every case performing far better than the isometric mutation operator. The mechanism for doing this, adjusting the range of parameter mutations, does not appear to scale well enough to handle very high dimensional problems.
    The reconstructive crossover operator is able to rapidly find high-performing solutions even in high-dimensional spaces, but is poor at exploring. Search with reconstructive crossover is confined to the distribution of genes that already exist in the archive, if used exclusively that distribution of genes is limited to the initial population.
    By combining these operators --- expanding the range of genes in the archive with mutation, and spreading high performing genes with reconstructive crossover --- DDE-Elites is able to create high-performing archives even in high-dimensional problems.


\subsection{Archive Recreation}
  DDE-Elites is as much a method of optimizing representations as solutions. By learning a representation from the archive, we create an encoding that is biased towards high performance and has a range of expression matching the defined behavior space. In these experiments, our DDE encodes smooth joint configurations which place an arm's end effector anywhere in its reach. To demonstrate that DDE-Elites does more than guide search, but learns a representation, we search the space again, using the found DDE in place of the direct encoding.

  We run the standard MAP-Elites algorithm, with isometric mutation only, using a learned DDE\footnote{The decoder network of the VAE found in the highest coverage replicate of DDE-Elites.} acting as our genome. In the 20D arm this DDE has 10 parameters, in the 200D and 1000D arms the DDE has 32 parameters. No previous solutions are maintained, \textit{only the trained DDE}. For reference we compare to the MAP-Elites algorithm using the direct encoding. An order of magnitude fewer evaluations were budgeted when using the DDE.

  In every case the DDE far outperforms the direct encoding, reaching the same levels of fitness and coverage with several orders of magnitude fewer evaluations (Figure~\ref{fig:recreate}). The DDE can express the same range of solutions as were found in the original archive, and finds them rapidly. Archives were recreated after only 10,000 evaluations --- a rate of about 5 evaluations per bin.\footnote{10,000 individuals/1950 bins $\approx$ 5 evaluations/bin discovered.}~The found solutions are also high performing.

  Such improvement cannot be explained away by the decrease in dimensionality of the search. In both low and high dimensional cases the bias toward high performance is also apparent: the mean fitness curve is nearly flat at the optima, indicating that when new solutions are added to the map they are already near optimal. The contrast with the direct encoding is stark, with the direct encoding considerable effort is taken to search for good solutions, the DDE finds little else. DDE-Elites not only produces solutions, but learns domain-specific representation.

\subsection{Optimization with Learned Encodings}
  Beyond its place in the DDE-Elites optimization loop, the produced DDE is a powerful representation with high expressivity and built in biases. Though created by MAP-Elites, the DDE is not tied to it. Once discovered, a DDE can be used as a representation for any black box optimization algorithm.

  We illustrate this generality by using again solving the arm inverse kinematics problem with the black-box optimizer CMA-ES~\cite{cmaes}. A set of target positions for the end effector is defined (Figure~\ref{fig:targetMatch}, left), and CMA-ES used to find a joint configuration which reaches each target. In one case optimization is performed using the DDE; in the other the direct encoding is used.

  When optimizing with the DDE, CMA-ES quickly finds solutions to the target hitting problems with a precision never matched with the direct encoding (Figure~\ref{fig:targetMatch}, top). Moreover, a bias for \textit{how} the problem is solved is built into the representation~(Figure~\ref{fig:targetMatch}, bottom). As the DDE was trained only on solutions with low joint variance, this same property is found in the solutions found by CMA-ES with the DDE --- even without searching for them. With the DDE CMA-ES not only finds solutions to the IK problem, the built-in priors of the DDE ensures we find kind of solutions we want.



\section{Discussion}
Learning representations by combining quality diversity (here, MAP-Elites) and generative models (here, a VAE) opens promising research avenues for domains in which optimizations of the same cost function are launched continuously. This is, for example, the case of Model Predictive Control \cite{mayne2000constrained}, in which the sequence of actions for the next seconds is optimized at every time-step of the control loop, or the case of shape optimization in interactive design tools \cite{hoyer2019neural,bendsoe1995optimization}, in which each modification by the user requires a novel optimization. 

In preliminary experiments, we searched for an encoding to describe action sequences for a walking robot. The results show that using MAP-Elites to generate a diversity of sequences, then using a VAE to learn a representation leads to an encoding that can accelerate future optimizations by several orders of magnitude. Nevertheless, using the representation during optimization, as described in this paper, did not accelerate the quality diversity optimization as much as in the high-dimensional arm used here. One hypothesis is that the regularities in action sequences are harder to recognize than in the arm experiments, especially at the beginning of the process. For instance, it might help to use an auto-encoder that is especially designed for sequences \cite{vaswani2017attention,co2018self}.

For other tasks, appropriate generative models could be explored, for example convolutional models for tasks with spatial correlations~\cite{salimans2015markov}. In addition, though the latent spaces created by VAEs are easier to navigate than those created by normal autoencoders, even better models offer the opportunities for further improvements. Much work has been done to create VAEs which have even better organized latent spaces~\cite{higgins2017beta,burgess2018understanding,chen2018isolating,kim2018disentangling}, ideally with each dimension responsible for a single phenotypic feature such as the lighting or color of an image.  

A second research avenue is to improve the bandit algorithm used to balance between operators. In theory, it should ensure that adding new operators can only aid optimization, since useless or detrimental operators would rarely be selected. However, we observed that it is not always effective: in some cases, using only the line mutation outperformed DDE-Elites, whereas DDE-Elites could revert to using only line mutation with a perfect bandit. Our hypothesis is that this is a sign that ``successes'' --- child solutions which discover new bins or improve on existing solutions --- is not the perfect measure of utility for a QD algorithm. In the case of our experiments, it may be that reconstructive crossover consistently improves solutions, but may only do so slightly. According to the ``success'' metric, a tiny improvement is worth the same as a large one. To best utilize the bandit, other methods of judging performance in QD algorithms should be explored.

Beyond performance advantages, for both the current and future optimizations, these ``disentangled'' representations offer even more interesting opportunities.
  Reducing the dimensionality of the search space into meaningful components would allow rapid model-based optimization of single solutions~\cite{bo}, or entire archives~\cite{gaier2018data}.
  Engineers could interactively explore and understand such encodings, laying bare the underlying properties responsible for performance and variation --- 
  %
  %
  and so from encodings receive, rather than provide, insight and domain knowledge.



 \section*{Acknowledgements}
 This work received funding from the European Research Council (ERC) under the EU Horizon 2020 research and innovation programme (grant agreement number 637972, project "ResiBots") and the German Federal Ministry of Education and Research (BMBF) under the Forschung an Fachhochschulen mit Unternehmen programme (grant agreement number 03FH012PX5, project "Aeromat").

\section*{Source Code}
 The source code used to produce the results in this paper is available at~\url{https://github.com/resibots/2020_gaier_gecco}

\bibliographystyle{style/ACM-Reference-Format}
\bibliography{bib_dde}

\clearpage

\onecolumn
\section*{Supplemental Material}
\renewcommand{\thesubsection}{\Alph{subsection}.}

\subsection{Example Maps}
\begin{table}[h!]
\begin{tabular}{ccc}
{\ul \textbf{Arm20}} & {\ul \textbf{Arm200}} & {\ul \textbf{Arm1000}} \\
\includegraphics[width=160pt]{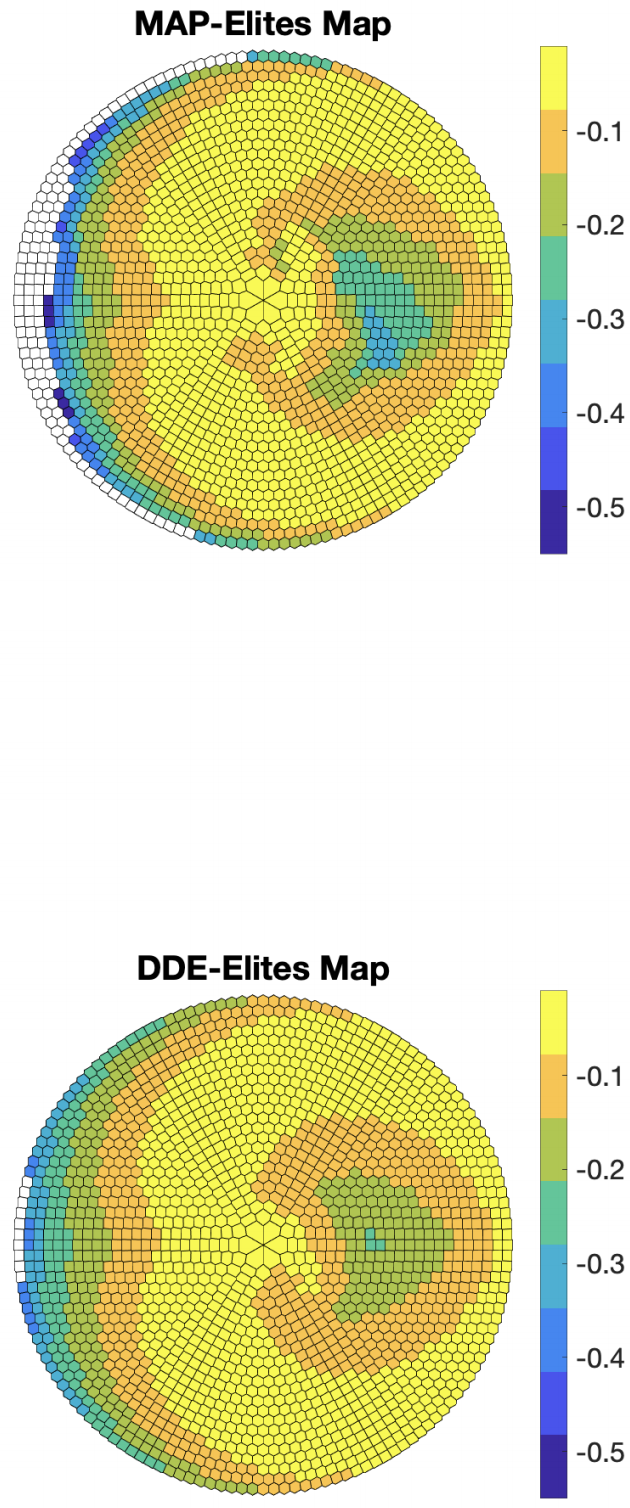} &
\includegraphics[width=160pt]{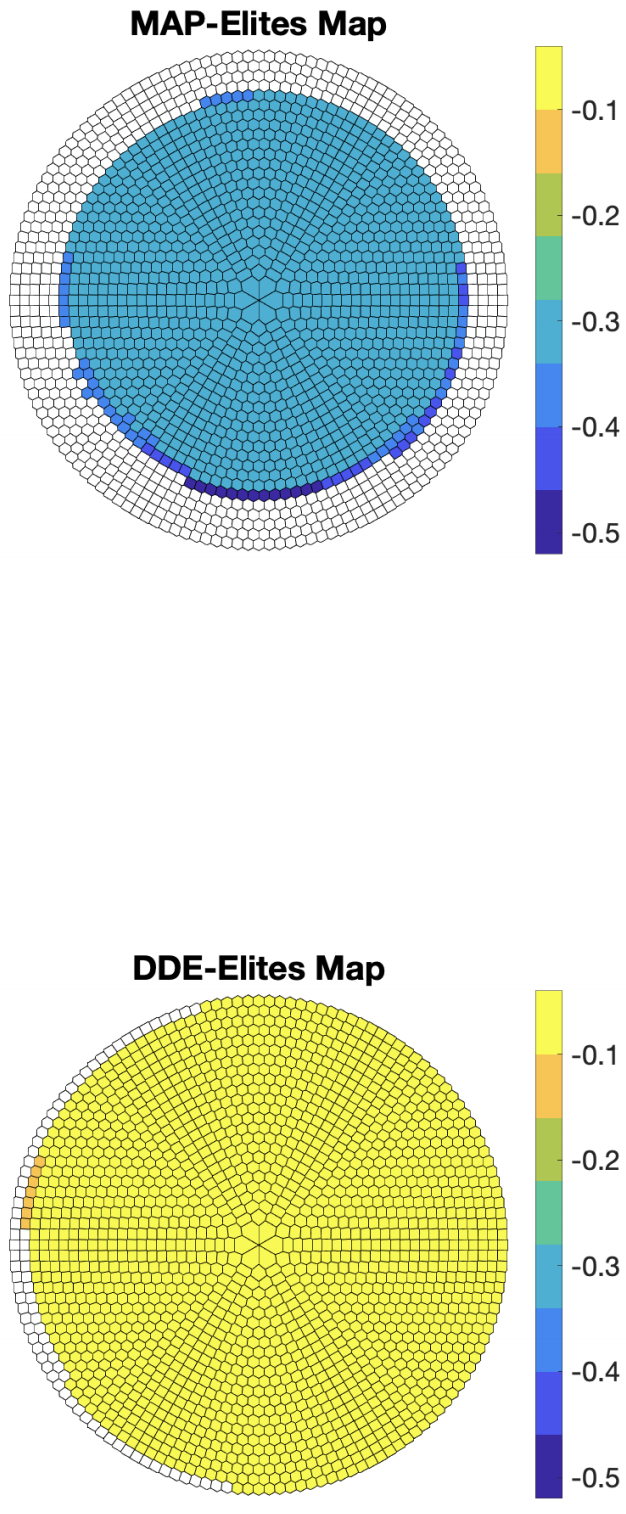} &
\includegraphics[width=160pt]{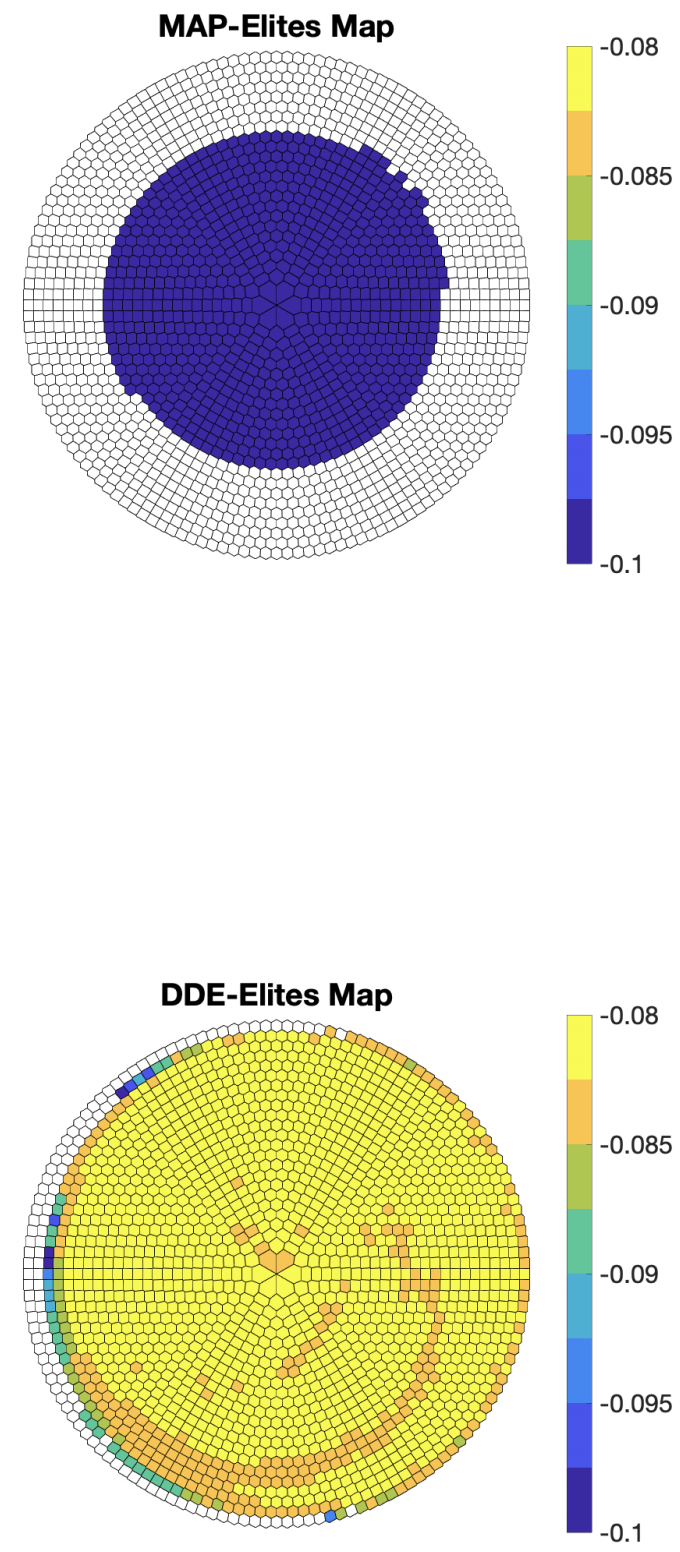}
\end{tabular}
\caption{
    \textit{Example Maps}
    \newline
    Final archives colored by fitness value for each cell for each domain. In the 20D Arm both MAP-Elites and DDE-Elites converge on similar optimal solutions. In the 200D and 1000D Arm MAP-Elites is unable to reach the levels of performance of DDE-Elites in any region.
}
\end{table}
\newpage

\subsection{Hyperparameters of DDE Experiments}
\begin{table}[h]
\begin{tabular}{|r|l|}
\hline
\textbf{Hyperparameter} & \textbf{Value} \\ \hline
Isometric Mutation Strength & 0.003 \\ \hline
Line Mutation Strength & 0.1 \\ \hline
Batch Size & 100 \\ \hline
Bandit Options, & \begin{tabular}[c]{@{}l@{}}{[}0.00:0.00:1.00{]}, {[}0.25:0.00:0.75{]},\\ {[}0.50:0.00:0.50{]}, {[}0.75:0.00:0.25{]},\\ {[}1.00:0.00:0.00{]}, {[}0.00:0.25:0.75{]},\\ {[}0.00:0.50:0.50{]}, {[}0.00:0.75:0.25{]},\\ {[}0.00:1.00:0.00{]}\end{tabular} \\ \hline
Bandit Window Length & 1000 \\ \hline
Generations per VAE Training & 1 \\ \hline
Epochs per VAE Training & 5 \\ \hline
Mutation Strength when Searching DDE & 0.15 \\ \hline
Latent Vector Length {[}Arm20{]} & 10 \\ \hline
Latent Vector Length {[}Arm200{]} & 32 \\ \hline
Latent Vector Length {[}Arm1000{]} & 32 \\ \hline
\end{tabular}
\end{table}

\end{document}